\documentclass[10pt,twocolumn,letterpaper]{article}

\usepackage{cvpr}              

\usepackage{amsthm}

\usepackage{wrapfig}  
\usepackage{graphicx}
\usepackage{tipa}
\usepackage{blindtext}
\usepackage{lineno}
\usepackage{colortbl}

\usepackage{amsmath,amssymb,amsfonts}
\usepackage{amsthm}
\usepackage{graphicx}

\usepackage{adjustbox}
\usepackage{wasysym}
\usepackage{footnote}
\usepackage{booktabs}
\usepackage{utfsym}
\usepackage{float}
\usepackage[dvipsnames]{xcolor}

\usepackage{colortbl}
\usepackage{soul}
\usepackage{algorithm}

\usepackage{url}
\usepackage[dvipsnames]{xcolor}
\usepackage[table]{xcolor}  
\usepackage{pifont}         
\definecolor{mydarkblue}{rgb}{0,0.08,0.45}
\definecolor{darkgreen}{rgb}{0.0, 0.5, 0.0} 
\definecolor{myblue}{RGB}{235,235,250}
\definecolor{lightblue}{RGB}{225, 235, 255}
\definecolor{lightgray}{RGB}{240, 240, 240}  
\definecolor{darkgray}{RGB}{220, 220, 220} 
\definecolor{superlightred}{rgb}{0.99, 0.92, 0.92}
\definecolor{darkgreen}{RGB}{50,100,0}
\definecolor{darkred}{RGB}{230, 150, 170}
\usepackage[most]{tcolorbox}

\usepackage{pgfplots}
\pgfplotsset{compat=1.17}
\usepackage{multirow}

\usepackage{algorithm}
\usepackage[noend]{algpseudocode}
\usepackage{amsmath,amssymb}
\usepackage{xcolor}
\algrenewcommand\algorithmiccomment[1]{\hfill \textcolor{gray}{// #1}}

\definecolor{uclablue}{RGB}{159, 195, 224}

\definecolor{uclagold}{RGB}{156, 172, 234}

\definecolor{grayred}{RGB}{30, 140, 224}

\usepackage{color, colortbl}

\usepackage[table]{xcolor}
\definecolor{cyan}{rgb}{0.573, 0.675, 0.878}
\definecolor{limegreen}{rgb}{0.675, 0.843, 0.557}

\definecolor{casegreen}{RGB}{117, 189, 67}

\newtcolorbox{ttcolorbox}[1][]{
  colframe=lightblue,
  colback=lightblue!5!white,
  title=#1,
  fonttitle=\bfseries\sffamily,
}

\usepackage[most]{tcolorbox}
\tcbuselibrary{breakable}

\newtcolorbox{remarkbox}[1][]{ 
  colback=uclagold!10,
  colframe=uclagold!80!black,
  coltitle=white,
  title=#1,
  fonttitle=\bfseries,
  colbacktitle=uclagold!80!black,
  left=6pt, right=6pt, top=6pt, bottom=6pt,
  breakable
}

\usepackage{amsthm}
\usepackage{mdframed}

\author{
    Zhongyu Yang\textsuperscript{\rm 1},
   Zuhao Yang\textsuperscript{\rm 2}, Shuo Zhan\textsuperscript{\rm 2}, Tan Yue\textsuperscript{\rm 3}, 
   Wei Pang\textsuperscript{\rm 1}, Yingfang Yuan\textsuperscript{\rm 1}\footnotemark[1] \\
  \textsuperscript{\rm 1}{BCML, Heriot-Watt University},\,    \textsuperscript{\rm 2}{Nanyang Technological University},\, \textsuperscript{\rm 3}{Peking University} \\ 
\parbox[c]{\textwidth}{\centering  
    \texttt{\{zy4028, y.yuan\}@hw.ac.uk}}}

\definecolor{cvprblue}{rgb}{0.21,0.49,0.74}
\usepackage[pagebackref,breaklinks,colorlinks]{hyperref}

\def\paperID{} 
\def\confName{CVPR}
\def\confYear{2026}

\title{SVAgent: Storyline-Guided Long Video Understanding \\via Cross-Modal Multi-Agent Collaboration}

\begin{document}

\twocolumn[{
\renewcommand\twocolumn[1][]{#1}
\vspace{-25pt}
\maketitle
\vspace{-20pt}
\begin{center}
\captionsetup{type=figure}
\centering
\includegraphics[width=\textwidth]{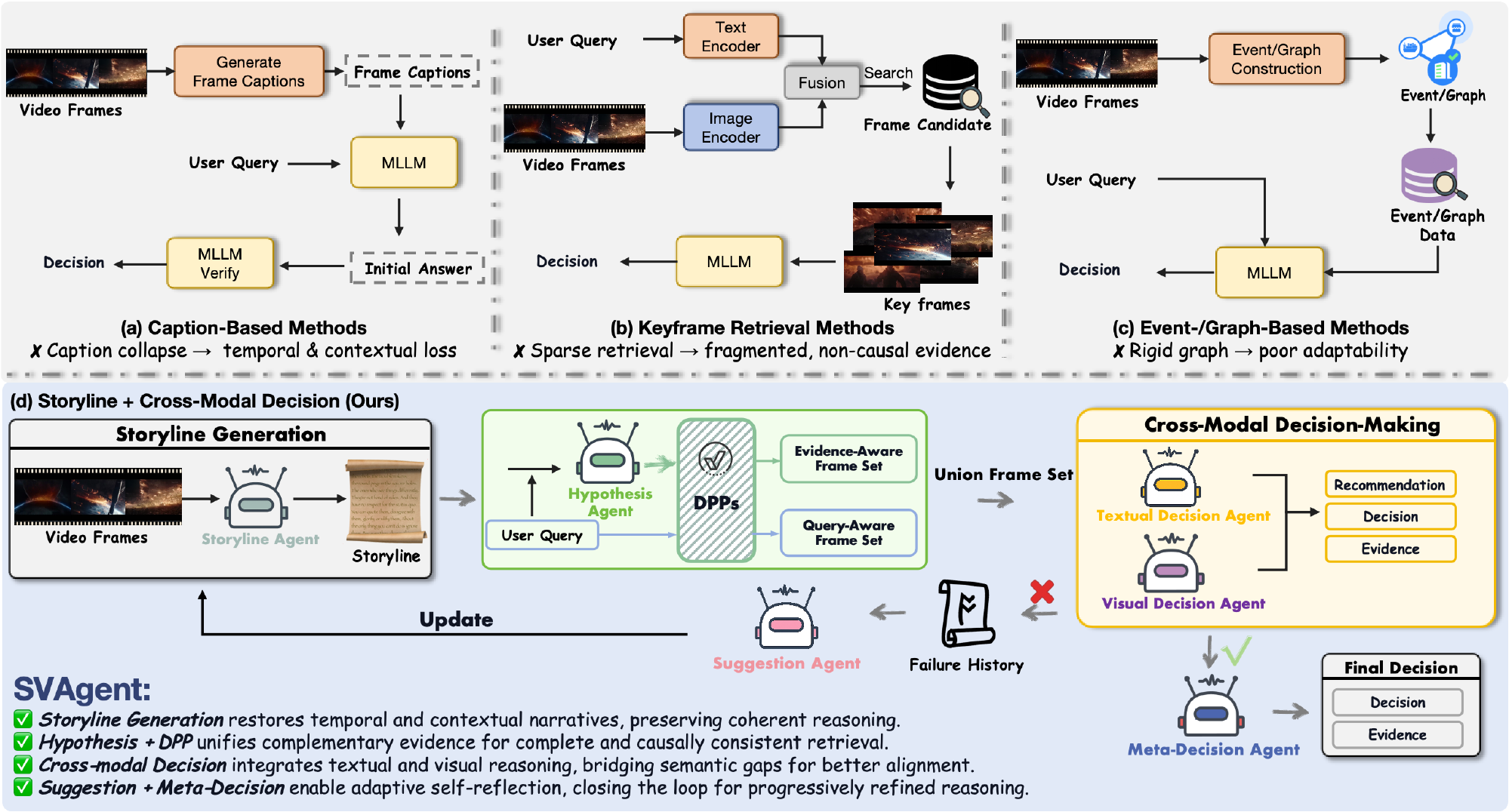}
  \caption{
  \textbf{Comparison of existing video reasoning paradigms and our proposed framework.} 
  (a) \textit{Caption-based} methods rely on text summaries, which often result in temporal collapse and insufficient visual grounding. 
  (b) \textit{Keyframe retrieval} methods focus on top-$k$ evidence, often leading to fragmented reasoning. 
  (c) \textit{Event/graph-based} approaches depend on fixed structures, which limits adaptability. 
  (d) We propose \textbf{SVAgent}, which unifies storyline reconstruction and cross-modal decision-making through multi-agent collaboration, recovers temporal narratives, ensures causal completeness via DPPs, and enables adaptive self-reflective reasoning beyond static paradigms.}
  \label{fig:teaser}
\end{center}
}]

\footnotetext[1]{$^*$ Corresponding author.}

\begin{abstract}
Video question answering (VideoQA) is a challenging task that requires integrating spatial, temporal, and semantic information to capture the complex dynamics of video sequences. Although recent advances have introduced various approaches for video understanding, most existing methods still rely on locating relevant frames to answer questions rather than reasoning through the evolving storyline as humans do. Humans naturally interpret videos through coherent storylines, an ability that is crucial for making robust and contextually grounded predictions. To address this gap, we propose SVAgent, a storyline-guided cross-modal multi-agent framework for VideoQA. The storyline agent progressively constructs a narrative representation based on frames suggested by a refinement suggestion agent that analyzes historical failures. In addition, cross-modal decision agents independently predict answers from visual and textual modalities under the guidance of the evolving storyline. Their outputs are then evaluated by a meta-agent to align cross-modal predictions and enhance reasoning robustness and answer consistency. Experimental results demonstrate that SVAgent achieves superior performance and interpretability by emulating human-like storyline reasoning in video understanding.
\end{abstract}
    
\section{Introduction}
\label{sec:intro}
Recent progress in multimodal AI has renewed interest in video understanding, which requires joint reasoning over visual content, language, and long-range temporal dynamics.
Compared with image-based visual question answering (VQA), VideoQA must reason over temporally dispersed evidence, making long-video understanding particularly challenging due to sparse key evidence and dominant irrelevant content.
Meanwhile, the large number of visual tokens in long videos can exceed the context window of current multimodal models, making exhaustive processing infeasible~\cite{liu2025shifting}.
This has driven recent progress in efficient visual processing for long-context multimodal models~\citep{liu2026globalcom2, liu2025mixkv, li2025todre, li2025comprehensive}, as well as in selective and hierarchical compression frameworks tailored to video and streaming video understanding~\citep{liu2025vidcom2, wang2025stc, script}.


Existing studies on video understanding generally fall into three representative paradigms (see  Figure~\ref{fig:teaser}).
\emph{Caption-based methods} generate textual descriptions for sparsely sampled frames and combine them with visual cues for downstream reasoning and answering~\citep{li2023videochat, cheng2024videollama,ataallah2024minigpt4}.
\emph{Keyframe retrieval methods} identify a compact set of query-relevant frames to support subsequent reasoning, reducing computational cost while retaining salient information~\citep{zhang2025qframe,wang2025videoitg}.
\emph{Event-/graph-based methods} explicitly construct higher-level temporal structures or relational graphs to facilitate structured and compositional reasoning~\citep{luo2023valley,shen2025vgent}.

Despite their progress, these paradigms remain limited in three aspects.
First, they lack an explicit mechanism for preserving global temporal structure. Caption-based methods model temporal relations only implicitly, while keyframe retrieval often disrupts temporal continuity, leading to fragmented reasoning.
Second, they provide limited support for trustworthy evidence acquisition, since locally selected frames or events may omit crucial context for reliable answering.
Third, they generally lack explicit verification, making them vulnerable to errors arising from ambiguous, inconsistent, or insufficient evidence.
Consequently, they struggle to jointly maintain temporal coherence, acquire reliable evidence, and ensure cross-modal consistency.

To address these limitations, we propose \textbf{SVAgent}, a multi-agent framework for long-video understanding inspired by how humans watch and reason about long videos.
Rather than reasoning over isolated observations, SVAgent first constructs an evolving storyline that serves as a coherent global temporal scaffold.
Based on this storyline, a hypothesis agent identifies query-relevant evidence, text, and vision agents perform complementary reasoning, and a meta-decision agent then explicitly verifies their cross-modal consistency.
When evidence remains insufficient or conflicting, a refinement suggestion agent proposes additional frames for targeted exploration, and the newly acquired observations are incorporated back into the storyline.

Overall, SVAgent thus forms a closed loop of global temporal modeling, selective evidence acquisition, cross-modal verification, and iterative refinement.

Our main contributions are as follows:
\begin{itemize}
    \item We propose \textbf{SVAgent}, a multi-agent framework that constructs and iteratively updates query-guided storylines to preserve global temporal structure.

    \item We introduce a verification-driven reasoning framework that couples evidence selection, cross-modal consistency checking, and targeted refinement to improve the reliability of long-video question answering.

    \item Experiments on four long-video benchmarks show consistent 5.5\%--11.5\% gains over baselines, demonstrating the effectiveness and robustness of SVAgent.
\end{itemize}
\begin{figure*}[ht]
\centering
\vspace{-5.5mm}
\includegraphics[width=\linewidth]{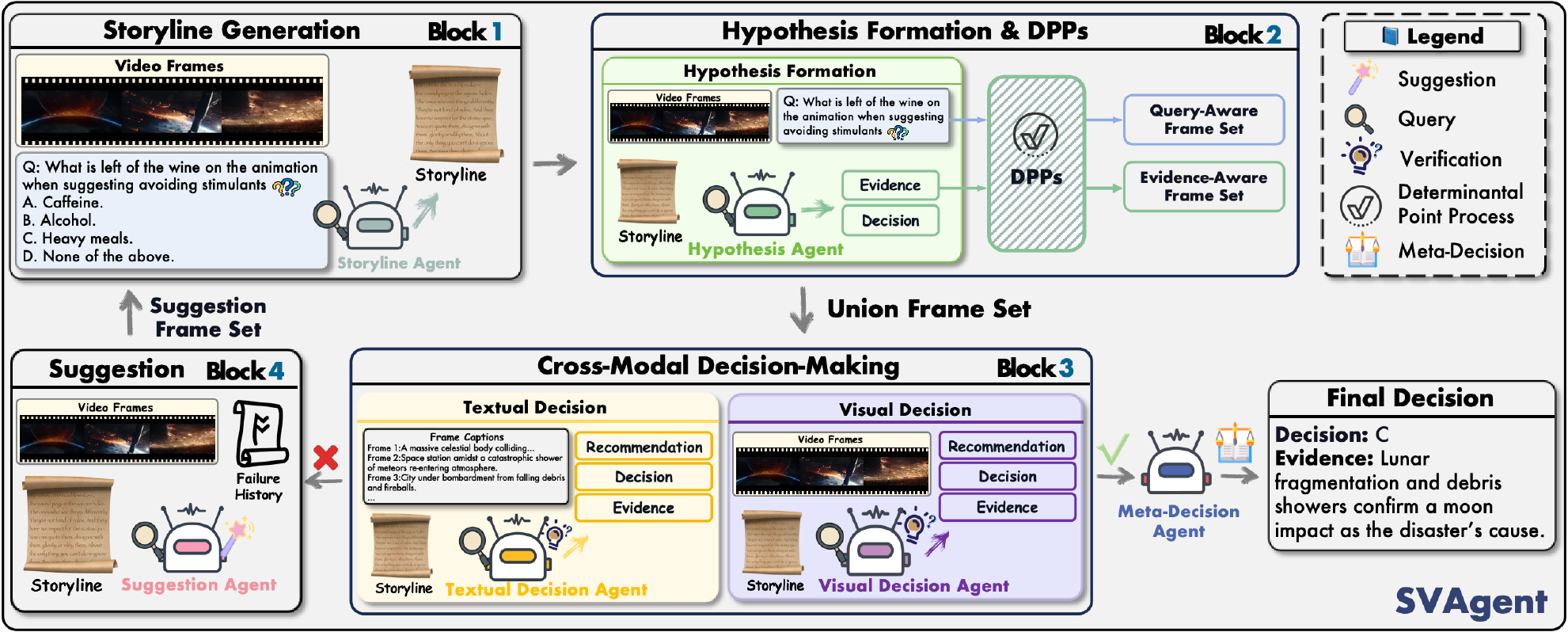}
\vspace{-5mm}
\caption{
Overview of \textbf{SVAgent}, which consists of six interacting agents:
(1) a \textbf{Storyline Agent} that summarizes sampled frames into a query-oriented narrative,
(2) a \textbf{Hypothesis Agent} that proposes and evaluates hypotheses while selecting evidence frames via Determinantal Point Processes (DPPs),
(3) two \textbf{Cross-Modal Decision Agents} that respectively conduct modality-specific reasoning to produce decisions, (4) a \textbf{Meta-Decision Agent} that verifies the consistency of cross-modal decisions,
and (5) a \textbf{Suggestion Agent} that uses error history to propose informative frames for iterative refinement.
The \textcolor{green!50!black}{\checkmark} indicates successful cross-modal decision confirmation,
while the \textcolor{red!70!black}{\texttimes} denotes inconsistency detection that triggers the refinement feedback loop.}
\label{fig:svagent}
\vspace{-5mm}
\end{figure*}

\section{Related Work}
\label{sec:related}

\subsection{Video Multimodal Large Language Models}
Multimodal Large Language Models (MLLMs)~\citep{chen2023minigpt,zhu2023minigpt,liu2024visual,liu2024llavanext,tong2024cambrian, zhang2025openmmreasoner} have been extended from images to videos, giving rise to Video MLLMs~\citep{cheng2024videollama,li2023videochat, li2024mvbench, Yang_2025_ICCV}.
Most approaches sample frames and encode them as visual tokens interleaved with text, providing a unified interface for image and video inputs~\citep{zhao2025humanomni,peng2025actionart,zhang2024video}.
However, in long videos, sparse evidence and long temporal spans make it difficult to preserve coherence and relevance under limited token budgets.

To address this, prior work has explored three main directions.
\emph{Caption-based methods}~\citep{li2023videochat,cheng2024videollama,ataallah2024minigpt4} compress video content into text for reasoning, but weaken visual grounding and capture temporal structure only implicitly.
\emph{Keyframe retrieval methods}~\citep{zhang2025qframe,wang2025videoitg,hu2025mllm,shen2024longvu} select relevant frames for reasoning, yet sparse selection breaks temporal continuity and fragments causal structure.
\emph{Event-/graph-based methods}~\citep{li2024mvbench,lin2023video,luo2023valley,shen2025vgent,li2025videochatr1} introduce structured representations, but depend on accurate evidence identification and remain less flexible for open-ended videos.
Overall, existing paradigms improve efficiency and representation but fail to explicitly model the interplay among temporal coherence, evidence selection, and reasoning, hindering consistent storylines and reliable evidence use in long videos.

\subsection{Agent-based Multimodal Reasoning}
Agentic reasoning has shown strong potential for long-horizon multimodal tasks, where iterative decomposition and exploration consistently improve over single-pass inference~\cite{wikiautogen, mermaid, yuan2025thinkvideosagenticlongvideo}.
Recent work further extends this paradigm with structured multi-agent coordination and self-reflective reasoning~\cite{zhang-etal-2024-omagent,zhang2025deep, XR, inex, yang2025longvt}.
In video understanding, this has inspired agent-based systems such as VideoAgent and its variants~\cite{fan2024videoagent,wang2024videoagent,videoagent2}, as well as hierarchical reasoning frameworks like VideoTree~\cite{ziyang2024videotree}.
Retrieval-augmented approaches further highlight the benefit of iterative querying and external evidence integration~\cite{ren2025videorag}.

\newpage
Despite these advances, several aspects remain underexplored.
Most approaches do not maintain an explicit query-guided storyline, making it harder to organize dispersed evidence into a coherent temporal interpretation.
Visual and textual signals are typically fused within a single decision pathway, limiting the ability to detect or resolve cross-modal inconsistencies.
In addition, refinement is often weakly guided, with limited mechanisms to identify which temporal segments should be revisited when evidence is insufficient or ambiguous~\cite{chen2025bring,yang2025vca,tang2025video}.

In contrast, SVAgent structures the reasoning process around a persistent storyline, explicit cross-modal decision making, and suggestion-driven refinement, all built on top of open Video MLLMs.
This results in a more controlled and interpretable reasoning process, where temporal coherence, evidence selection, and consistency checking are jointly modeled rather than handled in isolation.

\section{Method}
\label{sec:method}

SVAgent is a multi-agent system, as shown in Fig.~\ref{fig:svagent} and Algorithm~\ref{alg:svagent}, designed to decompose long video understanding tasks into four cooperative subtasks. 
SVAgent follows a cognitively inspired but modular design, where each agent is responsible for a distinct reasoning function and collectively enforces consistent cross-modal evidence validation.

SVAgent comprises four key components, each responsible for a specific stage of long-video reasoning.
\textbf{Storyline Generation} (Block~1) provides the foundation for subsequent reasoning.
A storyline serves as a compact temporal abstraction of the evolving video content and is progressively refined as new visual information becomes available.
\textbf{Hypothesis Formation and DPPs} (Block~2) then perform predictive reasoning by proposing answer hypotheses and identifying supporting or refuting evidence. The evidence and the query are used to retrieve frames via DPPs for verification. The ratio of the union of frame sets serves as coarse verification. When a hypothesis appears plausible, fine-grained verification is required for robustness.
Humans naturally cross-check information from different sources and re-examine whether the interpretation remains consistent. Motivated by this process, we introduce \textbf{Cross-modal Decision-Making} (Block~3), which validates conclusions by checking multimodal consistency and reinforcing relevant evidence. When the hypothesis test fails, the \textbf{Suggestion Agent} (Block~4) proposes additional informative frames to refine the understanding and support hypothesis updating, acting like a human who progressively skims a video based on previously reviewed content.

Overall, SVAgent integrates these components into a unified closed-loop, iterative framework that progressively refines its understanding of long videos through hypothesis-driven testing, cross-modal verification, and targeted exploration.
The following sections describe each component of the SVAgent framework in detail.

\begin{algorithm}[t]
\caption{Pseudocode of the SVAgent Framework}
\label{alg:svagent}
\begin{algorithmic}[1]
\Statex \textbf{Input:} Video frames $\mathcal{F}$, query $\mathcal{Q}$, options $\mathcal{O}$
\Statex \textbf{Output:} Prediction $o \in \mathcal{O}$

\State Initialize $\mathcal{C}, \mathcal{H}$ \Comment{\textcolor{gray}{captions, historical errors}}

\For{$n = 1$ to $T$}

    \If{$n = 1$}
        \State $\mathcal{C}', \mathcal{F}' \leftarrow \operatorname{Uniform\_Sampling}(\mathcal{C}, \mathcal{F})$
        \State $\mathcal{S} \leftarrow \mathcal{A}_s(\mathcal{C}', \mathcal{F}', \mathcal{Q})$ \\
        \Comment{\textcolor{blue!70!black}{storyline agent: initial storyline}}
    \Else
        \State $\mathcal{C}', \mathcal{F}' \leftarrow \mathcal{A}_r(\mathcal{H}, \mathcal{Q}, \mathcal{S})$ \\
        \Comment{\textcolor{teal!70!black}{refinement suggestion agent}}
        \State $\mathcal{S} \leftarrow \mathcal{A}_s(\mathcal{C}', \mathcal{F}', \mathcal{Q}, \mathcal{S})$ \\
        \Comment{\textcolor{blue!70!black}{storyline agent: storyline refinement}}
    \EndIf

    \State $\mathcal{E}, o \leftarrow \mathcal{A}_h(\mathcal{S}, \mathcal{Q}, \mathcal{F}')$ 
    \Comment{\textcolor{purple!70!black}{hypothesis agent}}

    \State $\mathcal{Y}_q \leftarrow \operatorname{DPP}(\mathcal{F}, \mathcal{Q})$
    \State $\mathcal{Y}_e \leftarrow \operatorname{DPP}(\mathcal{F}, \mathcal{E})$

    \If{$\dfrac{|\mathcal{Y}_q \cap \mathcal{Y}_e|}{|\mathcal{Y}_q| + |\mathcal{Y}_e|} > \alpha \ \operatorname{or}\ n = T$}
        \State $\mathcal{Y} \leftarrow \mathcal{Y}_q \cap \mathcal{Y}_e$
        \State $o_t, e_t, r_t \leftarrow \mathcal{A}_t(\mathcal{S}, \mathcal{C}_{\mathcal{Y}}, \mathcal{Q}, \mathcal{O})$ \\    
        \Comment{\textcolor{orange!85!black}{text-decision agent}}
        \State $o_v, e_v, r_v \leftarrow \mathcal{A}_v(\mathcal{S}, \mathcal{F}_{\mathcal{Y}}, \mathcal{Q}, \mathcal{O})$ \\
        \Comment{\textcolor{red!70!black}{vision-decision agent}}

        \If{$o_t = o_v \ \operatorname{or}\ n = N$}
            \State $o \leftarrow \mathcal{A}_m(o_t, e_t, r_t, o_v, e_v, r_v, \mathcal{S}, \mathcal{F}_{\mathcal{Y}})$ \\
            \Comment{\textcolor{magenta!70!black}{meta-decision agent}}
            \State \Return $o$
        \EndIf
    \EndIf

    \State Update $\mathcal{H}$
\EndFor
\end{algorithmic}
\end{algorithm}

\subsection{Storyline Generation}
The storyline serves as a compact, query-aware abstraction of the video, retaining reasoning-relevant temporal cues while suppressing redundant frame-level content.
Instead of treating frames as isolated observations, it models their temporal and semantic dependencies, forming a coherent narrative scaffold for subsequent inference.
Given a video and its query, we convert the video into a sampled frame set $\mathcal{F}$ and generate corresponding captions $\mathcal{C}$.
The storyline agent $\mathcal{A}_s$ then produces a query-conditioned storyline
$\mathcal{S} \leftarrow \mathcal{A}_s(\mathcal{C}', \mathcal{F}', \mathcal{Q})$,
where $\mathcal{F}'$ and $\mathcal{C}'$ denote the currently selected frame–caption subset.
The design of $\mathcal{A}_s$ follows three principles.
First, the agent incorporates the query $\mathcal{Q}$ to focus on reasoning-relevant temporal and semantic content, reducing the search space by suppressing irrelevant segments.
Second, as full video processing is often unnecessary in practical VideoQA scenarios, the agent operates on sampled frames $\mathcal{F}'$ rather than the full sequence.
As shown in Algorithm~\ref{alg:svagent} (Lines~4 and~7), these frames are obtained either from initial uniform sampling or suggestion-driven sampling, the latter provided by the suggestion agent.
Once sufficient evidence is gathered, the process naturally terminates.
Third, the storyline supports progressive refinement, updating its abstraction as new frames become available and revising incomplete or uncertain narrative segments.

In this way, $\mathcal{A}_s$ maintains a globally coherent yet adaptive storyline that integrates available evidence, infers missing temporal links, and progressively refines its abstraction to support the question-answering objective.

\subsection{Hypothesis Formation and DPPs}
The hypothesis stage provides the predictive reasoning and testing mechanisms of SVAgent.
This process follows an exploration–exploitation pattern. The current storyline and sampled frames guide hypothesis formation, which is then verified using selectively retrieved evidence. To support this process, we introduce a hypothesis agent coupled with a DPP-based mechanism for adaptive evidence selection.

\begin{table*}[!ht]
\resizebox{\linewidth}{!}{
\begin{tabular}{lccccccccc}
\toprule  
\multirow{2}{*}{\textbf{Model}} & \multirow{2}{*}{\textbf{Size}} & \multirow{2}{*}{\textbf{\# Frames}} &
\textbf{LongVideoBench} & 
\textbf{MLVU} & 
\textbf{LVBench} & 
\multicolumn{4}{c}{\textbf{VideoMME (w/o vs. w/ sub.)}} \\ 
\cmidrule(lr){4-4} \cmidrule(lr){5-5} \cmidrule(lr){6-6} \cmidrule(lr){7-10} 
&&& \textit{\textbf{val (w/o sub.)}} & \textit{\textbf{test}} & \textit{\textbf{test}} & \textit{\textbf{short}} & \textit{\textbf{medium}} & \textit{\textbf{long}} & \textit{\textbf{overall}} \\
\midrule
\multicolumn{10}{c}{\textbf{\textit{Proprietary Video MLLMs}}} \\
\midrule
\rowcolor{gray!10} Gemini 1.5 Pro~\citep{team2024gemini}  & - & 1fps & 64.0 & 64.0 & 33.1 & 81.7 / 84.5 & 74.3 / 81.0 & 67.4 / 77.4 & 75.0 / 81.3  \\
\rowcolor{gray!10} GPT-4o~\citep{openai2024gpt4o} & - & 1fps  & 66.7 & 64.6 & 48.9 & 80.0 / 82.8 & 70.3 / 76.6 & 65.3 / 72.1 & 71.9 / 77.2 \\
\midrule
\multicolumn{10}{c}{\textbf{\textit{Open-source Video MLLMs}}} \\
\midrule
LLaVA-Video~\cite{zhang2025llavavideo} & 72B & 32 / 64\textsuperscript{$\ast$} & 61.9 & 72.9 & 38.7 & 80.7 / 81.8  & 68.7 / 73.8 & 62.1 / 72.2 & 70.3 / 75.9 \\
Qwen2.5-VL~\citep{bai2025qwen2} & 72B & 32 / 768\textsuperscript{$\ast$} & 62.5 & 73.1 & 39.3 & 79.4 / - & 71.2 / - & 71.8 / - & 74.1 / 79.1 \\
InternVL 2.5~\citep{chen2024expanding}  & 78B & 64 & 63.6 & 75.7 & 43.6 & 82.8 / - & 70.9 / - & 62.6 / - & 72.1 / 74.0\\
\midrule
\multicolumn{10}{c}{\textbf{\textit{Open-source Video Agents}}} \\
\midrule
VideoMind~\cite{liu2025videomind} & 7B & 32 & 56.3 & 64.4 & 40.8 & - / - & - / - & 49.2 / - & 58.2 / -  \\
Vgent~\cite{shen2025vgent} & 7B & 320 & 59.7 & 72.1 & - & - / - & - / - & - / - & 68.9 / 74.3  \\
Video-RAG~\cite{videorag} & 72B & 64 & 65.4 & 73.8 & -  & 82.8 / - & 76.3 / - & 73.1 / - & 77.4 / - \\
\midrule
\multicolumn{10}{c}{\textbf{\textit{Baseline vs. Ours}}} \\
\midrule
Qwen2.5-VL~\citep{bai2025qwen2} & 3B & 8 & 53.0 & 53.6 & 31.6 & 63.7 / 67.3 & 51.6 / 62.0 & 43.1 / 50.6 & 52.8 / 60.0 \\
\rowcolor{lightblue} 
+ \textbf{SVAgent (Ours)} & 3B & 8 & 59.7\textcolor{blue}{\textsuperscript{+6.7}} & 61.2\textcolor{blue}{\textsuperscript{+7.6}} & 38.5\textcolor{blue}{\textsuperscript{+6.9}} & 71.0\textcolor{blue}{\textsuperscript{+7.3}} / 72.4\textcolor{blue}{\textsuperscript{+5.1}} & 59.4\textcolor{blue}{\textsuperscript{+7.8}} / 66.3\textcolor{blue}{\textsuperscript{+4.3}} & 51.6\textcolor{blue}{\textsuperscript{+8.5}} / 58.3\textcolor{blue}{\textsuperscript{+7.7}} & 60.7\textcolor{blue}{\textsuperscript{+7.9}} / 65.7\textcolor{blue}{\textsuperscript{+5.7}} \\
Qwen3-VL~\citep{qwen3} & 4B & 8 & 52.5 & 53.4 & 32.1 & 64.5 / 69.6 & 50.6 / 61.4 & 46.6 / 52.2 & 53.9 / 61.1  \\
\rowcolor{lightblue} 
+ \textbf{SVAgent (Ours)} & 4B & 8 & 57.0\textcolor{blue}{\textsuperscript{+11.5}} & 64.9\textcolor{blue}{\textsuperscript{+7.3}} & 37.6\textcolor{blue}{\textsuperscript{+5.5}} & 69.7\textcolor{blue}{\textsuperscript{+5.2}} / 74.5\textcolor{blue}{\textsuperscript{+4.9}} & 57.1\textcolor{blue}{\textsuperscript{+6.5}} / 66.6\textcolor{blue}{\textsuperscript{+5.2}} & 53.5\textcolor{blue}{\textsuperscript{+6.9}} / 57.5\textcolor{blue}{\textsuperscript{+5.3}} & 60.1\textcolor{blue}{\textsuperscript{+6.2}} / 66.2\textcolor{blue}{\textsuperscript{+5.1}} \\
\midrule
Qwen2.5-VL~\citep{bai2025qwen2} & 7B & 8 & 54.8 & 54.7  & 33.9 & 63.9 / 69.0 &  50.2/ 62.0 & 44.4 / 52.3 & 52.8 / 61.1  \\
\rowcolor{lightblue} 
+ \textbf{SVAgent (Ours)} & 7B & 8 & 60.7\textcolor{blue}{\textsuperscript{+5.9}} & 62.7\textcolor{blue}{\textsuperscript{+9.3}} & 40.6\textcolor{blue}{\textsuperscript{+6.7}} & 72.1\textcolor{blue}{\textsuperscript{+8.2}} / 75.8\textcolor{blue}{\textsuperscript{+6.8}} & 58.0\textcolor{blue}{\textsuperscript{+7.8}} / 67.5\textcolor{blue}{\textsuperscript{+5.5}} & 53.3\textcolor{blue}{\textsuperscript{+8.9}} / 57.3\textcolor{blue}{\textsuperscript{+5.0}} & 61.2\textcolor{blue}{\textsuperscript{+8.3}} / 66.9\textcolor{blue}{\textsuperscript{+5.8}} \\
Qwen3-VL~\citep{qwen3} & 8B & 8 & 54.9 & 54.5 & 33.4 & 66.7 / 71.7 & 53.4 / 65.2& 47.4 / 54.6 & 55.8 / 63.8  \\
\rowcolor{lightblue} 
+ \textbf{SVAgent (Ours)} & 8B & 8 & 61.0\textcolor{blue}{\textsuperscript{+6.1}} & 65.6\textcolor{blue}{\textsuperscript{+11.1}} & 40.8\textcolor{blue}{\textsuperscript{+7.4}} & 73.9\textcolor{blue}{\textsuperscript{+7.2}} / 78.0\textcolor{blue}{\textsuperscript{+6.3}} & 61.5\textcolor{blue}{\textsuperscript{+8.1}} / 71.1\textcolor{blue}{\textsuperscript{+5.9}} & 53.9\textcolor{blue}{\textsuperscript{+6.5}} / 60.5\textcolor{blue}{\textsuperscript{+5.9}} & 63.1\textcolor{blue}{\textsuperscript{+7.3}} / 69.8\textcolor{blue}{\textsuperscript{+6.0}} \\
\bottomrule
\end{tabular}
}
\caption{\textbf{Comparison on Long Video Understanding Benchmarks.}}
\label{tab:main}
\end{table*}

\textbf{Hypothesis Agent} $\mathcal{A}_h$ is responsible for generating a tentative answer hypothesis and identifying its supporting evidence.
Given the storyline $\mathcal{S}$, sampled frames $\mathcal{F}'$, query $\mathcal{Q}$, and answer options $\mathcal{O}$, the agent predicts the most plausible answer and extracts a corresponding evidence set $\mathcal{E}$ from the available visual and textual cues.
The extracted evidence is used in later stages to inform subsequent frame selection and test the hypothesis.

\textbf{Determinantal Point Processes (DPPs)} are widely used in video and text retrieval to preserve salient content while avoiding redundant selections. In SVAgent, two DPPs are employed, conditioned on the query $\mathcal{Q}$ and the evidence $\mathcal{E}$, to retrieve key frame subsets from the full video. A plausible hypothesis leads to strong agreement between the two selections, indicating consistent and relevant evidence. 
To this end, we construct corresponding DPP kernels, where the kernel matrix for each DPP is defined as:
\begin{equation}
    L=\operatorname{diag}(\mathbf{r}) S \operatorname{diag}(\mathbf{r}),
\end{equation}
where $S$ denotes the similarity matrix for all frames, and $\mathbf{r}$ represents the relevance vector. 

For the query-conditioned DPP, $\mathbf{r}_q$ measures query–frame relevance via textual–visual similarity to align selected frames with the question.
The second DPP focuses on evidence, modeling how well frames support or refute a hypothesis; accordingly, $\mathbf{r}_e$ quantifies frame–evidence relevance. Together, they provide complementary query- and evidence-driven signals. The two kernels independently select $k$ key frames using a fast greedy MAP algorithm~\cite{chen2018fast}. The resulting sets are combined via union and intersection to capture complementary coverage and shared evidence, serving as a test of how well the hypothesis is formed. If the intersection ratio, defined as $|\mathcal{Y}_q \cap \mathcal{Y}_e|/k$, exceeds a threshold $\alpha$, indicating sufficient agreement, the process proceeds to cross-modal decision and verification; otherwise, the suggestion agent $\mathcal{A}_r$ proposes additional frames for refinement. 

This design treats $\mathcal{Q}$ as an anchor to ensure that selected frames remain closely tied to the query, while $\mathcal{E}$ advances the reasoning process toward potential answers. The intersection ratio (Line 12) thus reflects semantic consistency between the two selections: a low ratio indicates weak alignment between question- and evidence-driven cues, signaling the need for further exploration.

\subsection{Cross-Modal Decision and Verification}
In human decision-making, after forming and testing a hypothesis, we often seek extra verification to increase confidence in the result. This may involve checking multiple sources or using peer review to ensure agreement. Inspired by this behavior, we design a Cross-Modal Decision-Making process that works after hypothesis testing to improve the reliability of the final answer. It includes a Text Decision Agent and a Vision Decision Agent, followed by a Meta-Decision Agent. The first two provide opinions from different modalities, and the meta-agent focuses on reaching an agreement to make the final decision.

\textbf{Text and Vision Decision Agents} $\mathcal{A}_t$ and $\mathcal{A}_v$ denote the text- and vision-based decision agents, respectively.
The text agent $\mathcal{A}_t$ takes $\mathcal{S}$, $\mathcal{Q}$, and the captions of the union frame set $\mathcal{C}_Y$ as input, while the vision agent $\mathcal{A}_v$ operates on the corresponding frames $\mathcal{F}_Y$ together with $\mathcal{S}$ and $\mathcal{Q}$.
In this way, the two agents independently leverage modality-specific information to produce complementary judgments.
The hypothesis decision $o$ is not explicitly included, as its information is assumed to be implicitly propagated through the frame selection stage.
This shared representation enables $\mathcal{A}_t$ and $\mathcal{A}_v$ to produce reliable decisions while reducing error propagation from explicit information leakage.

\textbf{Meta-Decision Agent}
Once the two decision agents produce the same decision, we introduce a meta-decision agent $\mathcal{A}_m$ to verify the result. $\mathcal{A}_m$ receives three types of information from the decision agents $\mathcal{A}_t$ and $\mathcal{A}_v$: the decisions $o_t$ and $o_v$, the evidence $e_t$ and $e_v$ derived from the observed information used during decision making such as the storyline, captions, or frames, and the recommendations $r_t$ and $r_v$ that indicate the importance of frames. To integrate these signals, $\mathcal{A}_m$ evaluates the consistency between $o_t$ and $o_v$ and examines whether the corresponding evidence $e_t$ and $e_v$ mutually support or contradict each other. If the two decisions differed in the final iteration, $\mathcal{A}_m$ performs a reconciliation step by weighting the evidence with the frame-importance recommendations $r_t$ and $r_v$, enabling it to favor the decision supported by stronger or more reliable observations. If the decisions agree, $\mathcal{A}_m$ conducts a secondary verification by cross-checking the evidence to ensure that the agreement is not accidental or based on incomplete information. This meta-decision mechanism provides a robust layer of cross-modal validation, reduces error propagation, and improves the reliability of the final answer.

\subsection{Suggestion Agent}
As discussed earlier, human video understanding unfolds as an iterative process. In quick VideoQA scenarios, we naturally skim through the video and skip over segments that appear uninformative, focusing instead on regions that seem more relevant to the question. This selective behavior is guided by reasoning, as humans continually predict where missing or clarifying information is most likely to appear.

Inspired by this intuition, we design the Suggestion Agent $\mathcal{A}_r$, whose goal is to determine which additional frames should be consulted when the current evidence is insufficient for reliable decision making. After each unsuccessful attempt, $\mathcal{A}_r$ examines the indices of the frames that have already been used for storyline construction and decision making. These frames represent the agent’s current understanding of the video, but also reveal which portions of the video have not contributed enough information.

Building on this observation, $\mathcal{A}_r$ operates as a targeted retrieval policy that prioritizes regions with high uncertainty and potential relevance to the query $\mathcal{Q}$.
Specifically, it identifies (1) temporal regions that remain unexplored or weakly informative, and (2) regions likely to contain evidence aligned with the query.
Based on these signals, $\mathcal{A}_r$ infers a set of candidate frames that maximize expected information gain and reduce residual uncertainty.
Rather than uniform or heuristic sampling, the agent performs reasoning-driven selection, adaptively focusing on frames that are most informative for hypothesis refinement.
This design enables efficient and targeted exploration of long videos, where evidence is sparse and unevenly distributed.
\section{Experiment}

\noindent \textbf{Baselines and Models.} 
We evaluate SVAgent on available Video MLLMs, primarily Qwen2.5-VL~\cite{bai2025qwen2} and Qwen3-VL~\cite{Qwen3-VL}, spanning model sizes from 3B to 8B.
We further compare against strong long-video baselines, including \textbf{Videomind}~\citep{liu2025videomind}, \textbf{Video-RAG}~\citep{luo2024video}, and \textbf{VideoAgent}~\citep{wang2024videoagent}. 

\noindent \textbf{Benchmarks.}
We evaluate SVAgent across four representative long-video benchmarks:
\begin{itemize}
    \item \textbf{Video-MME}~\citep{fu2024video} is a widely used evaluation suite that spans 11 seconds to 1 hour and assesses detailed real-world long-video comprehension.
    \item \textbf{MLVU}~\citep{zhou2024mlvu} contains videos ranging from 3 minutes to 2 hours (avg 12 min) and focuses on multi-stage events and long-range temporal dependencies.
    \item \textbf{LongVideoBench}~\citep{wu2025longvideobench} targets referred and multi-hop reasoning, where questions require analysing long frame sequences rather than isolated frames.
    \item  \textbf{LVBench}~\citep{wang2024lvbench} emphasizes long-range temporal grounding and multi-event reasoning, with videos spanning several minutes to over an hour.
\end{itemize}

Across all benchmarks, questions depend on extended temporal context and cannot be reliably answered using single-frame or sparsely sampled observations.

\noindent \textbf{Implementation Details:}
For storyline construction, each video is sampled at 1.0 FPS to form a frame database, from which 10\% of frames are uniformly selected as the initial pool.
Similarity computation in DPPs uses \texttt{google/siglip-so400m-patch14-384}~\cite{siglip}.
The intersection ratio is set to $\tau=0.3$, and SVAgent runs for a maximum of 3 refinement iterations.
All experiments are executed on NVIDIA H100 80GB GPUs.
The same hyperparameters are used for all benchmarks.

\subsection{Main Results}
Table~\ref{tab:main} summarizes performance across four long-video benchmarks. Baseline video MLLMs often bias toward early frames or visually salient yet incomplete cues, while SVAgent redirects retrieval toward temporally relevant segments, reducing temporal bias, local over-reliance, and modality drift. Across all benchmarks and model sizes, SVAgent yields consistent gains, with larger improvements on longer videos that require extended reasoning.

On \textbf{LongVideoBench}, which emphasizes multi-hop reasoning, SVAgent improves Qwen2.5-VL-3B from 53.0 to 59.7 and Qwen3-VL-4B from 52.5 to 57.0. These +6.7 and +4.5 improvements show that stabilizing the temporal narrative is especially beneficial for smaller models, which otherwise miss later events.
A similar trend appears on \textbf{MLVU}. Multi-stage events distributed across long intervals often cause baselines to favor visually prominent but semantically incomplete evidence. SVAgent improves Qwen2.5-VL-3B from 53.6 to 61.2 and Qwen3-VL-4B from 53.4 to 64.9, with gains of +7.6 and +11.5, respectively, confirming that anchoring selection to the evolving storyline enables DPPs to retrieve more temporally aligned frames.
On \textbf{LVBench}, where long-range grounding is crucial, SVAgent provides consistent improvements of roughly +6 points across all model sizes. This reflects its robustness against short-term correlations that frequently mislead baselines in repetitive or visually similar scenes.
The effect is clearest on \textbf{VideoMME}. Longer videos amplify modality drift in baselines, causing textual and visual predictions to diverge. SVAgent mitigates this effect, improving Qwen2.5-VL-7B from 69.0 to 75.8 on medium videos (+6.8) and from 52.3 to 57.3 on long videos (+5.0), as the refinement loop repeatedly realigns modalities to storyline-consistent evidence.

\begin{table}[]
\resizebox{\linewidth}{!}{
\begin{tabular}{cccc cc ccc}
\toprule
\multirow{2}{*}{\textbf{Story}} & 
\multicolumn{2}{c}{\textbf{Verification}} & \multirow{2}{*}{\textbf{Meta}}  & 
\textbf{LongVideoBench}& 
\textbf{MLVU} & 
\textbf{LVBench}  & 
\multicolumn{2}{c}{\textbf{VideoMME}} \\ 
\cmidrule(lr){2-3} \cmidrule(lr){5-5} \cmidrule(lr){6-6} \cmidrule(lr){7-7} \cmidrule(lr){8-9} 
& \textbf{Textual} & \textbf{Visual} & & \textit{\textbf{val (w/o sub.)}} & \textit{\textbf{test}} & \textit{\textbf{test}} & \textit{\textbf{long}} & \textit{\textbf{overall}} \\
\midrule
\textcolor{red}{\usym{1F5F6}} & \textcolor{red}{\usym{1F5F6}} & \textcolor{red}{\usym{1F5F6}} & \textcolor{red}{\usym{1F5F6}} & 54.8 & 54.7 & 33.7 & 44.4 / 52.3 & 52.8 / 61.1\\
\textcolor{green}{\usym{2714}} & \textcolor{red}{\usym{1F5F6}} & \textcolor{red}{\usym{1F5F6}} & \textcolor{red}{\usym{1F5F6}} & 56.1 & 57.9 & 37.4 & 47.9 / 54.8 & 55.6 / 63.5 \\
\textcolor{green}{\usym{2714}} & \textcolor{green}{\usym{2714}} & \textcolor{red}{\usym{1F5F6}} & \textcolor{red}{\usym{1F5F6}} & 56.8 & 58.3 & 36.9 & 49.1 / 53.9 & 57.3 / 62.5 \\
\textcolor{green}{\usym{2714}} & \textcolor{red}{\usym{1F5F6}} & \textcolor{green}{\usym{2714}} & \textcolor{red}{\usym{1F5F6}} & 59.1 & 61.3 & \underline{39.7} & \underline{50.1} / 55.9 & 58.1 / 64.5\\
\textcolor{red}{\usym{1F5F6}} & \textcolor{green}{\usym{2714}} & \textcolor{green}{\usym{2714}} & \textcolor{red}{\usym{1F5F6}} & 58.2 & 58.1 & 38.9 & 49.7 / 53.1 & 59.4 / 64.3 \\
\textcolor{green}{\usym{2714}}  & \textcolor{green}{\usym{2714}} & \textcolor{green}{\usym{2714}} & \textcolor{red}{\usym{1F5F6}} & \underline{59.3} & \underline{62.5} & 39.1 & 48.2 / \underline{56.3} & \underline{59.9} / \underline{65.3}\\
\rowcolor{lightblue}
\textcolor{green}{\usym{2714}} & \textcolor{green}{\usym{2714}} & \textcolor{green}{\usym{2714}} & \textcolor{green}{\usym{2714}} & \textbf{60.7} & \textbf{62.7} & \textbf{40.6} & \textbf{53.3 / 57.3} & \textbf{61.2 / 66.9} \\
\bottomrule
\end{tabular}
}
\caption{\textbf{Ablation Studies on Qwen2.5VL-7B.}}
\label{tab:ablation}
\vspace{-3mm}
\end{table}

\subsection{Ablation Studies}
\noindent \textbf{Key components.} Table~\ref{tab:ablation} reports ablations on Qwen2.5-VL-7B. Removing the storyline agent yields the lowest scores across all benchmarks, often dropping performance by 4--8 points, highlighting that without an explicit temporal scaffold, the model struggles when key events are spread across long spans. Enabling the storyline agent alone produces consistent gains such as +5.6 on LongVideoBench and +4.9 on LVBench, where reconstructing multi-hop or referred events is essential.
Adding textual or visual verification further improves performance, and the relative gains differ by benchmark. Textual checks contribute more to MLVU (+3.8), especially for questions that require fine-grained semantic alignment, while visual checks help on VideoMME long videos (+3.5) by filtering out visually inconsistent frames. Using both verifiers together strengthens cross-modal consistency.
The meta-decision agent provides the final and largest boost, adding +3.0 on LongVideoBench and +2.6 on MLVU by resolving conflicts between textual and visual branches, especially when evidence is sparse or ambiguous. 
Overall, the ablation trends indicate that each component reinforces a different aspect of temporal or modal reliability, and their integration yields the full robustness of SVAgent.

\begin{figure}[!t]
\centering
\includegraphics[width=\linewidth]{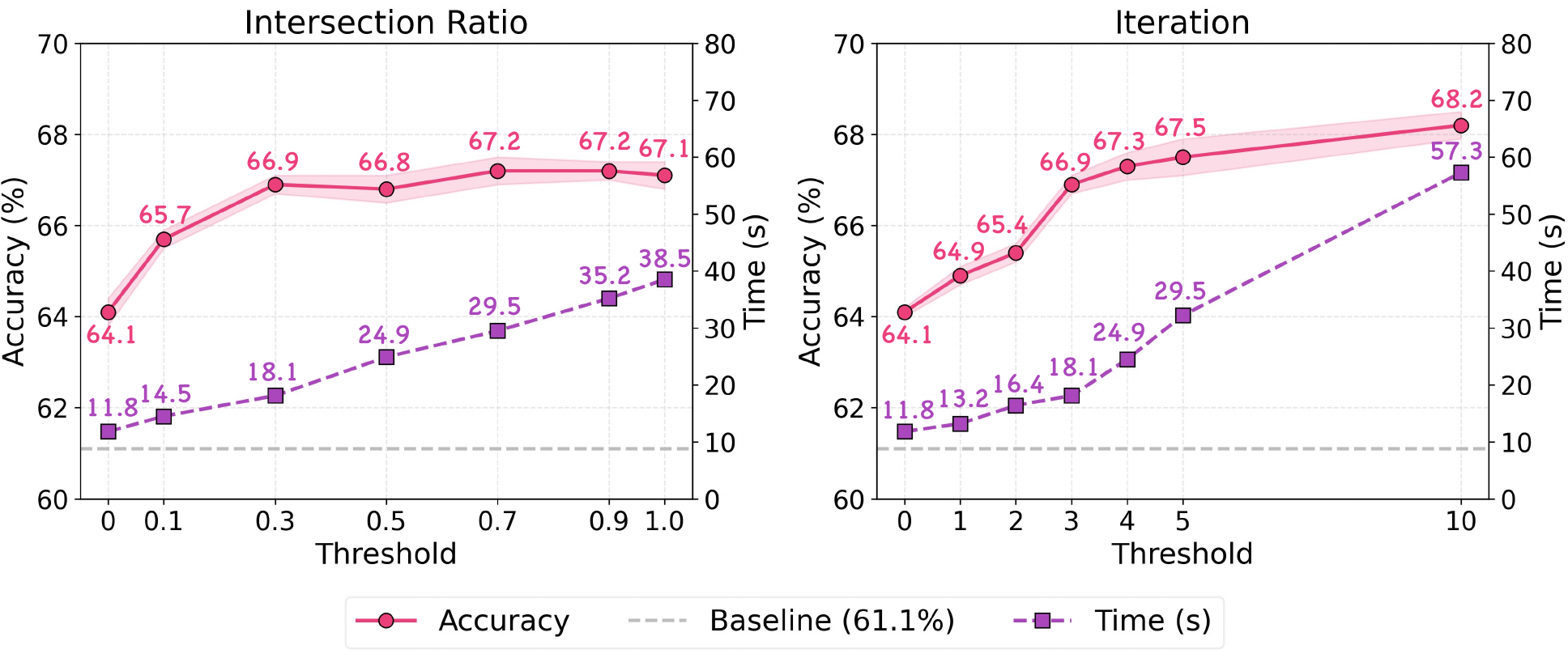}
\vspace{-3mm}
\caption{Parameter and Latency Analysis on Video-MME.}
\label{fig:para}
\end{figure}

\noindent \textbf{Retrieval models.}
\begin{table}[]
\resizebox{\linewidth}{!}{
\begin{tabular}{c cc ccc}
\toprule
\multirow{2}{*}{\textbf{Retrieval Model}}  & 
\textbf{LongVideoBench}& 
\textbf{MLVU} & 
\textbf{LVBench}  & 
\multicolumn{2}{c}{\textbf{VideoMME}} \\ 
\cmidrule(lr){2-2} \cmidrule(lr){3-3} \cmidrule(lr){4-4} \cmidrule(lr){5-6}
& \textit{\textbf{val (w/o sub.)}} & \textit{\textbf{test}} & \textit{\textbf{test}} & \textit{\textbf{long}} & \textit{\textbf{overall}} \\
\midrule
CLIP~\cite{clip} & \underline{60.2} & 61.9  &  \underline{40.4} & \textbf{54.2} / \underline{56.6}  &  60.7 / 65.2 \\
LongCLIP~\cite{longclip} & 59.9 &  \textbf{63.1} & 39.8 & 52.5 / 55.1  &  60.4 / 64.9 \\
\rowcolor{lightblue}
Siglip & \textbf{60.7} & \underline{62.7} & \textbf{40.6} & \underline{53.3} / \textbf{57.3} & \textbf{61.2 / 66.9} \\
\bottomrule
\end{tabular}
}
\caption{\textbf{Ablation study on the retrieval model in DPP.}}
\label{tab:retrival}
\end{table}

Table~\ref{tab:retrival} compares different encoders used in the DPPs retrieval module.  
While CLIP, LongCLIP, and SigLIP vary in their visual–text alignment properties, the performance gap across benchmarks remains small, and all retrieval models enable SVAgent to maintain consistent accuracy.  
This pattern shows that once a reasonable relevance signal is provided, SVAgent’s iterative reasoning and verification pipeline consistently refines and stabilizes the evidence, making the system largely insensitive to the specific choice of retrieval backbone.  
These results indicate that SVAgent’s performance depends more on downstream reasoning mechanisms than on the initial encoder used for frame relevance estimation.

\subsection{Parameter and Latency Analysis}
Figure~\ref{fig:para} analyzes how the intersection ratio $\tau$ and refinement iteration count $T$ affect accuracy and runtime. Increasing $\tau$ raises the agreement requirement between query-driven and evidence-driven DPPs selections, which improves accuracy from 64.1 to 67.2 as the selected frames become more semantically aligned with the storyline. This stricter filtering, however, increases latency from 11.8s to 38.5s due to the higher cost of computing cross-set consistency over a reduced but more concentrated frame pool. 
Adjusting $T$ shows a complementary trend: additional refinement cycles yield accuracy gains up to 68.2, with latency remaining moderate for $T{=}1$–$5$ (11.8s to 24.9s) before rising at $T{=}10$ (57.3s) as repeated DPPs and meta-decision steps accumulate. 
Overall, $\tau$ governs the strictness of evidence intersection and $T$ controls the depth of iterative reasoning, and $\tau{=}0.3$, $T{=}3$ emerges as an effective balance between accuracy and computational cost.

\begin{figure}[!t]
\centering
\includegraphics[width=\linewidth]{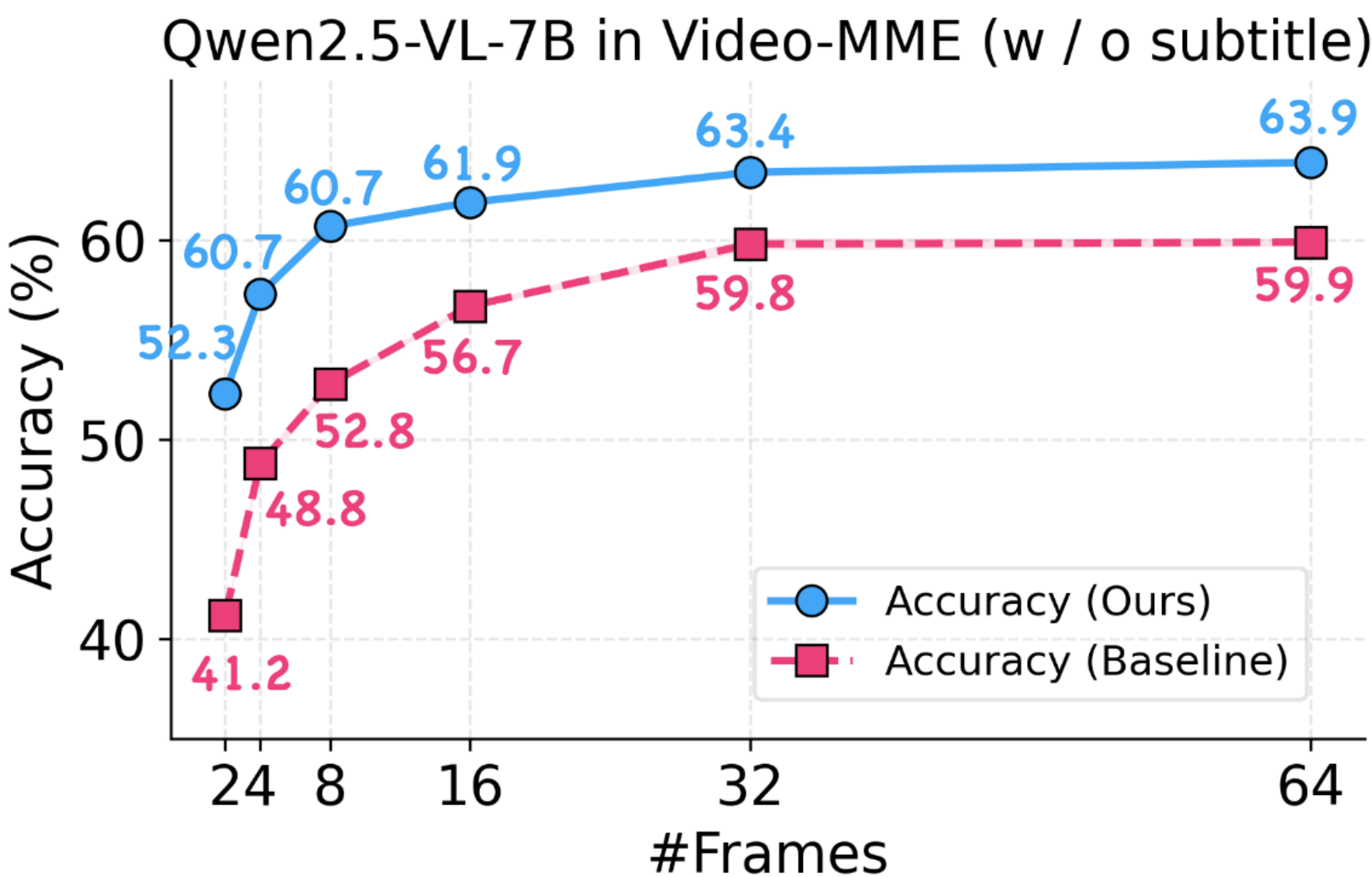}
\vspace{-3mm}
\caption{Comparison of SVAgent and uniform sampling under different frame budgets on Video-MME.}
\label{fig:frame_selection}

\end{figure}

\begin{figure*}[ht]
\centering
\includegraphics[width=\linewidth]{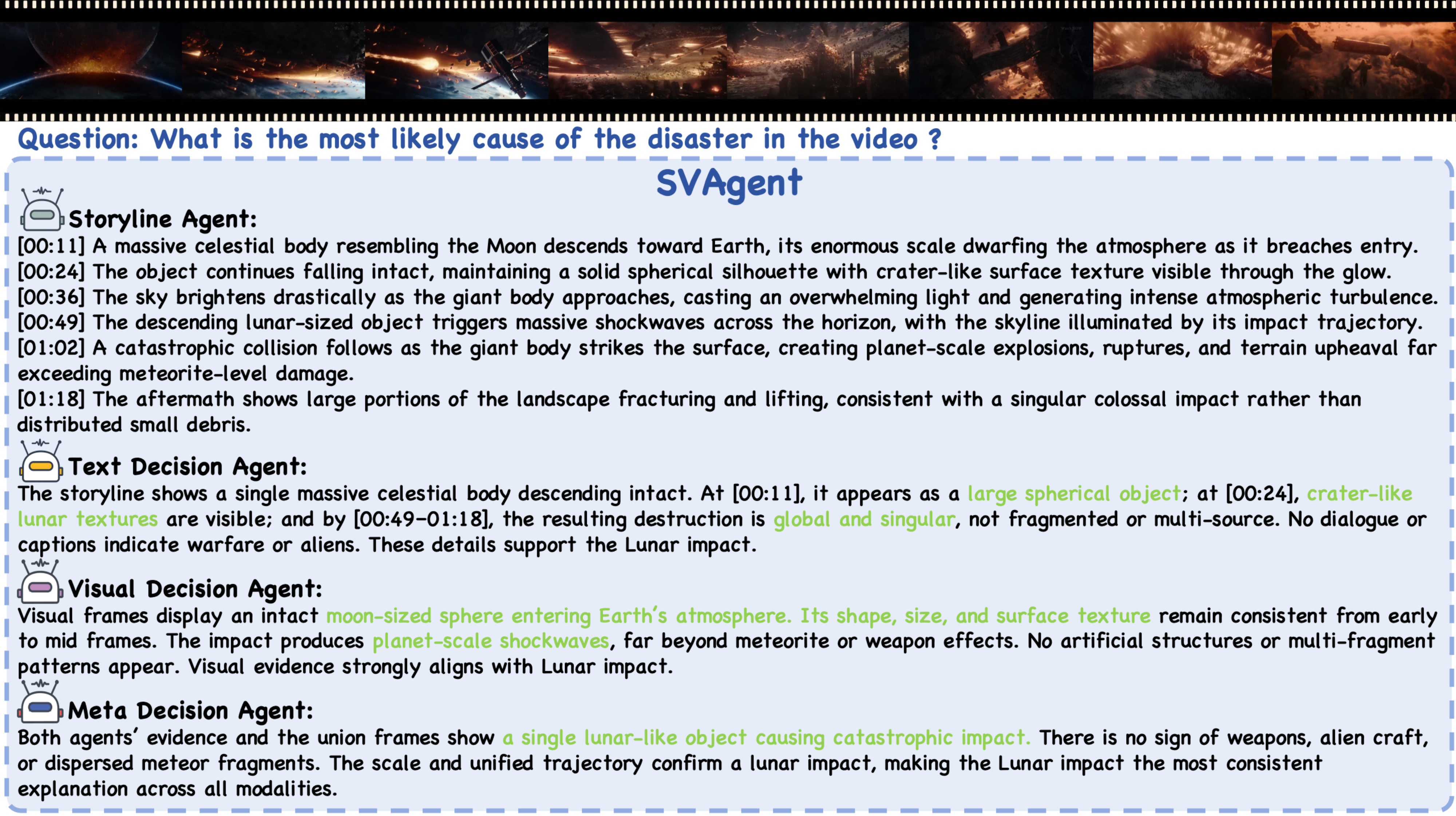}
\vspace{-2mm}
\caption{Case study showing how SVAgent maintains coherent temporal reasoning and consistent cross-modal verification. 
By combining storyline-guided abstraction with complementary text and vision decisions, the system retrieves stable evidence across long sequences and converges on reliable explanations for complex video events.}
\label{fig:case}
\end{figure*}

\subsection{Frame Selection Analysis}
Figure~\ref{fig:frame_selection} compares uniform sampling with SVAgent under different frame budgets. 
Uniform sampling improves quickly at small budgets, rising from 41.2\% at 4 frames to 59.8\% at 32 frames, but its gains taper off as additional frames become increasingly redundant. 
In contrast, SVAgent achieves 60.7\% accuracy at only 8 frames and maintains a clear margin at every setting, reaching 63.9\% at 64 frames. 
These trends indicate that long-video reasoning benefits more from selecting informative and query-relevant frames than from increasing frame quantity, enabling SVAgent to deliver reliable performance even when operating with a limited subset of the video.

\subsection{Statistical Significance Study}
\begin{table}[!t]
\centering
\resizebox{\linewidth}{!}{
\begin{tabular}{lcccc}
\toprule
\textbf{Method} & \textbf{Mean (\%)} & \textbf{StdDev} & \textbf{$t$-test $p$} & \textbf{Wilcoxon $p$}  \\
\midrule
Raw            & 54.68 & 0.89 & \multirow{2}{*}{$7.09\times10^{-10}$} & \multirow{2}{*}{$9.77\times10^{-4}$} \\
\textbf{SVAgent} & \textbf{62.76} & \textbf{0.31} &  &  \\
\bottomrule
\end{tabular}
}
\caption{
Statistical comparison on the \textsc{MLVU} benchmark. 
\textbf{StdDev} denotes standard deviation. 
We report one-sided paired $t$-test and Wilcoxon signed-rank test $p$-values ($\alpha = 5\%$). 
Both tests show that \textbf{SVAgent significantly outperforms Raw}.}
\label{tab:significance}
\end{table}

To quantitatively assess the robustness of the performance improvements, we conduct 10 independent runs with distinct random seeds under identical evaluation settings. We then apply paired one-sided $t$-tests and Wilcoxon signed-rank tests to determine whether the gains achieved by SVAgent are statistically significant. The null hypothesis ($\mathbf{H}_{0}$) assumes no improvement over the Raw baseline, whereas the alternative hypothesis ($\mathbf{H}_{1}$) states that SVAgent achieves superior accuracy.
As reported in Table~\ref{tab:significance}, SVAgent attains a higher mean score of 62.76 with a standard deviation of 0.31, compared to 54.68 with a standard deviation of 0.89 for the Raw baseline, indicating both improved accuracy and reduced variability. The resulting $p$-values from the $t$-test ($7.09 \times 10^{-10}$) and Wilcoxon test ($9.77 \times 10^{-4}$) are far below the significance threshold $\alpha = 0.05$, leading to a rejection of $\mathbf{H}_{0}$. These findings provide strong statistical evidence that SVAgent consistently outperforms the Raw baseline across repeated trials.

\subsection{Case Study}
Figure~\ref{fig:case} shows an example in which SVAgent analyzes a long sequence depicting a lunar-sized object approaching Earth. The storyline agent organizes the sampled frames into a clear narrative that captures the object’s descent and the scale of its impact. The text and vision decision agents then highlight complementary cues, such as the object’s intact spherical shape and the global extent of the resulting destruction. The meta-decision agent integrates these signals and identifies a singular large-body impact as the most consistent explanation. Beyond this specific case, the behavior observed here reflects a broader pattern in which SVAgent maintains stable evidence selection and coherent reasoning over extended temporal spans, enabling reliable interpretations even in visually complex long-video scenarios.

\section{Conclusion}
In this work, we present SVAgent, a storyline-guided multi-agent framework for long-video understanding. By coupling query-conditioned storylines with iterative verification, SVAgent recovers temporally distributed evidence, enabling reliable reasoning when key observations are scattered across distant segments. Experiments on four long-video benchmarks show consistent gains across backbone scales and video durations, especially on tasks requiring long-range temporal composition, multi-hop inference, and cross-modal evidence coordination. Ablations further show that these gains arise from the complementary roles of the agents in temporal grounding, semantic disambiguation, and verification. Overall, our experimental results highlight structured, query-driven, and iterative evidence aggregation as a foundation for long-video reasoning.



\newpage
{
    \small
    \bibliographystyle{ieeenat_fullname}
    \bibliography{main}
}

\end{document}